\documentclass[10pt,twocolumn,letterpaper]{article}

\usepackage{cvpr}
\usepackage{times}
\usepackage{epsfig}
\usepackage{graphicx}
\usepackage{amsmath}
\usepackage{amssymb}
\usepackage{dsfont}
\usepackage{booktabs}
\usepackage{siunitx}
\usepackage{bm}
\usepackage{dblfloatfix}
\usepackage{xcolor}

\usepackage[pagebackref=true,breaklinks=true,letterpaper=true,colorlinks,bookmarks=false]{hyperref}

\definecolor{selected_concept_color}{HTML}{666666}
\definecolor{independent_concept_color}{HTML}{6C8EBF}
\definecolor{common_concept_color}{HTML}{B85450}

\cvprfinalcopy 

\def\cvprPaperID{3448} 

\ifcvprfinal\fi
\begin{document}

\title{Learning Unsupervised Visual Grounding\\Through Semantic Self-Supervision}

\author{Syed Ashar Javed\\
The Robotics Institute\\
Carnegie Mellon University\\
{\tt\small sajaved@andrew.cmu.edu}
\and
Shreyas Saxena\\
{\tt\small shreyas.saxena2@gmail.com}
\and
Vineet Gandhi\\
Kohli Center of Intelligent Systems (KCIS)\\
IIIT-Hyderabad\\
{\tt\small vgandhi@iiit.ac.in}
}

\maketitle

\begin{abstract}
Localizing natural language phrases in images is a challenging problem that requires joint understanding of both the textual and visual modalities. In the unsupervised setting, lack of supervisory signals exacerbate this difficulty. In this paper, we propose a novel framework for unsupervised visual grounding which uses concept learning as a proxy task to obtain self-supervision. The intuition behind this idea is to encourage the model to localize to regions which can explain some semantic property in the data, in our case, the property being the presence of a concept in a set of images. We present thorough quantitative and qualitative experiments to demonstrate the efficacy of our approach and show a $5.6\%$ improvement over the current state of the art on Visual Genome dataset, a $5.8\%$ improvement on the ReferItGame dataset and comparable to state-of-art performance on the Flickr30k dataset. 
\end{abstract}

\section{Introduction}
The recent advancements in computer vision have seen the problem of visual localization evolve from using pre-defined object vocabularies, to arbitrary nouns and attributes, to the more general problem of grounding arbitrary length phrases. Utilizing phrases for visual grounding overcomes the limitation of using a restricted set of categories and provides a more detailed description of the region of interest as compared to single-word nouns or attributes.
%
%

Recent works have used supervised learning for the task of visual grounding (i.e localizing)~\cite{fukui2016multimodal,plummer2017phrase,Chen_2017_ICCV, rohrbach2016grounding}. However, these approaches require expensive bounding box annotations for the phrase, which are difficult to scale since they are a function of scene context and grow exponentially with the number of entities present in the scene. Furthermore, bounding box annotations for phrases are subjective in nature and might contain non-relevant regions with respect to the phrase. This brings us to our main motivation, which is to explore new ways in which models can directly harness unlabelled data and its regularities to learn visual grounding of phrases.
%
\begin{figure}[t]
\centering
\includegraphics[width=\linewidth]{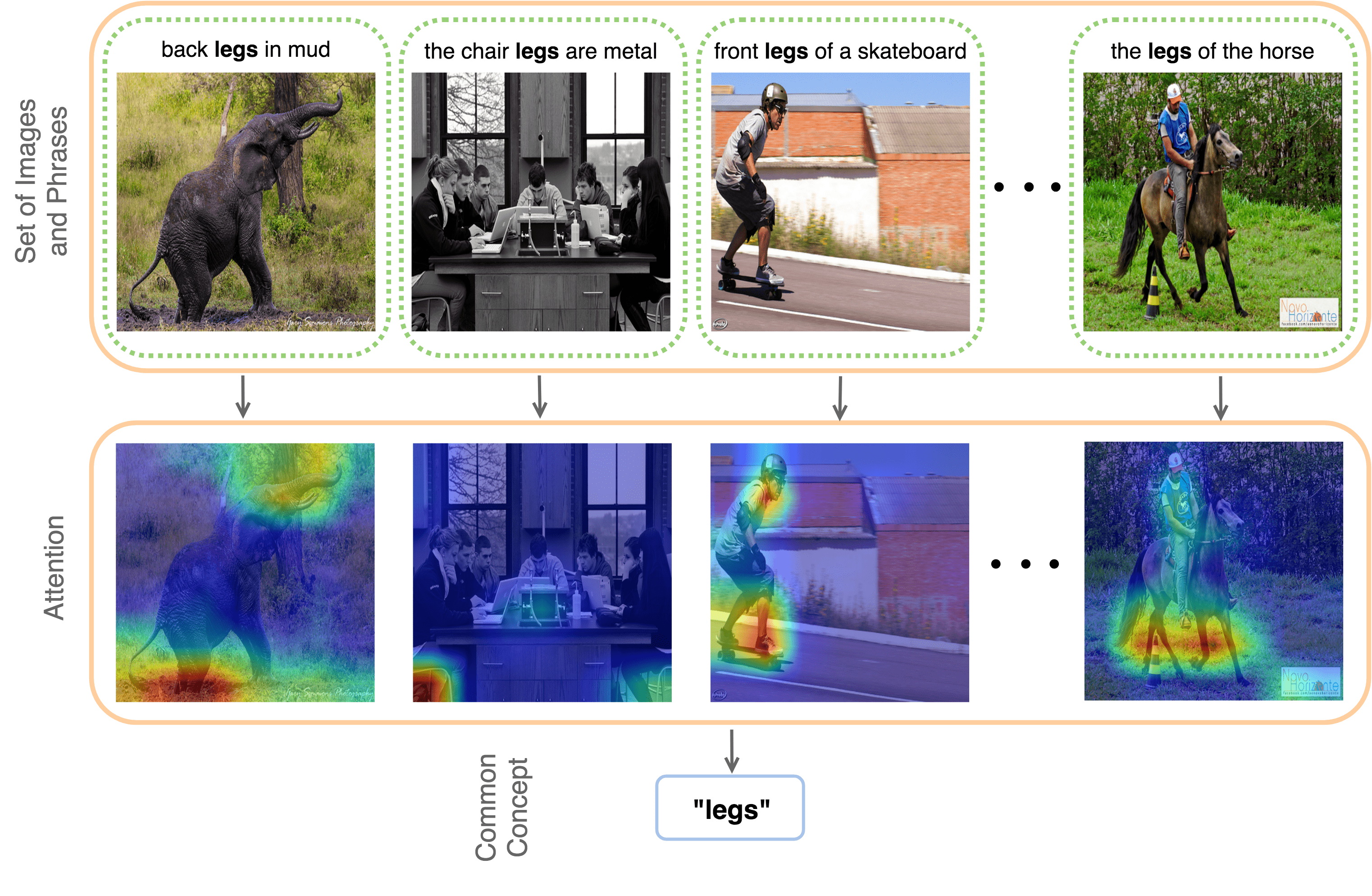}
\caption{We exploit the presence of semantic commonalities within a set of image-phrase pairs to generate supervisory signals. We hypothesize that to predict these commonalities, the model must localize them correctly within each image of the set.}
\label{fig:motivation}
\end{figure}

Given the lack of supervision, we develop a self-supervised proxy task which can be used for guiding the learning. The general idea behind self-supervision is to design a proxy task which involves explaining some regularity about the input data. Since there are no ground truth annotations, the model is trained with a surrogate loss which tries to optimize for a proxy task, instead of directly optimizing for the final task. A good proxy task improves performance on the final task when the surrogate loss is minimized.

In this work we propose concept-learning as a substitute task for visual grounding. During training, we create \emph{concept batches} of size $k$, consisting of $k$ different phrase-image pairs, all containing a common concept (as illustrated in Figure~\ref{fig:motivation}). The proxy task for the model is to decode the common concept present within each concept batch. We induce a parametrization which, given the input text and image, can generate an attention map to localize a region. These localized regions are then used to predict the common concept. Adopting concept-learning as our substitute task, we align our proxy and empirical task, and by introducing concept batches, we constrain the model to learn concept representations across multiple contexts in an unsupervised way.

Previous work on unsupervised visual grounding can also be interpreted as having proxy losses to guide the localization. ~\cite{rohrbach2016grounding} use reconstruction of the whole phrase as a substitute task for grounding. However, we hypothesize that the objective of reconstructing the entire phrase can also be optimized by learning co-occurrence statistics of words and may not always be a result of attending to the correct bounding box. Moreover, precise reconstruction of certain uninformative parts of the phrase might not necessarily correlate well with the correct grounding. This limitation is also evident in other methods like that of~\cite{Xiao_2017_CVPR} which uses a discriminative loss on the whole phrase instead of generating discrimination for the object to be localized. Many other works like~\cite{Ramanishka2017cvpr} and~\cite{zhang2016top} only allow for word-level grounding, thus making them average over the heatmaps to get a phrase-level output. In contrast, our formulation does not suffer from these limitations. Our proxy task deals with the full phrase and forces the model to limit the attention to areas which can explain the concept to be grounded, thus aligning the objective better with the task of visual grounding.

To evaluate the generality of our approach, we test our approach on three diverse datasets. Our ablations and analysis identify certain trends which highlight the benefits of our approach. In summary, the main contributions of our work are as follows:
\begin{itemize}
\item We propose a novel framework for visual grounding of phrases through semantic self-supervision where the proxy task is formulated as concept learning. We introduce the idea of a concept batch to aid learning.
\item We evaluate our approach on the Visual Genome and ReferIt dataset and achieve state-of-art performance with a gain of $5.6\%$ and $5.8\%$ respectively. We also get performance comparable to the state-of-art on Flickr30k dataset.
\item We analyze the behavior of our surrogate loss and the concept batch through thorough ablations which gives an insight into the functioning of our approach. We also analyze the correlation of performance for visual grounding with respect to size of the bounding box and possible bias induced due to the similarity of the grounded concepts to the ImageNet labels.
\end{itemize}
\section{Related work}
\textbf{Visual grounding of phrases.} The problem of image-text alignment has received much attention in the vision community in the recent years. Early work like DeViSE~\cite{frome2013devise} focus on learning semantic visual embeddings which have a high similarity score with single-word labels. Similar to DeViSE,~\cite{Ren_etal_BMVC_17} learn a multi-modal alignment by constructing a semantic embedding space, but instead of image-label correspondences, they learn region-label correspondences through a multiple-instance learning approach. 
~\cite{kiros2014unifying} learn a joint embedding space for a complete sentence and an image using a CNN-LSTM based encoder and a neural language model based decoder.
Since the release of the Flickr30k Entities dataset~\cite{plummer2015flickr30k} and subsequently the Visual Genome dataset~\cite{krishna2017visual}, availability of bounding box annotations of phrases has allowed many new attempts at the problem of visual grounding of phrases.~\cite{plummer2015flickr30k} provide a baseline for Flickr30k Entities dataset using Canonical Correlation Analysis (CCA) to compute the region-phrase similarity.~\cite{wang2016learning} construct a two-branch architecture that enforces a structure and bi-directional ranking constraint to improve upon the CCA baseline.
Another recent work from~\cite{Chen_2017_ICCV} departs from the standard usage of bounding box proposals and uses the primary entity of the phrase along with its context to regress localization coordinates. They use a combination of a regression, a classification and a reinforcement learning based loss to train multiple networks in their framework. 
Prior to our work, there are two papers which take up the problem of unsupervised visual grounding of phrases.~\cite{rohrbach2016grounding} use reconstruction of the original phrase as a substitute objective function to improve visual attention. But the output predictions in their work is in the form of bounding boxes which, as noted by~\cite{Chen_2017_ICCV}, puts an upper bound on the performance. In a more recent work,~\cite{Xiao_2017_CVPR} use the parent-child-sibling structure in the dependency tree of the phrase along with a discriminative loss to generate weak supervision and produce heatmap based outputs for localization. Following~\cite{Xiao_2017_CVPR}, we too generate heatmap based localizations, but use an objective which is better aligned with the grounding task. Apart from these papers, certain other unsupervised methods allow modification of their approach to enable evaluation on the phrase grounding task. For example,~\cite{zhang2016top} and~\cite{Ramanishka2017cvpr} produce word-level heatmaps and average them to get a phrase-level output. We compare with all these works in section~\ref{sec:result_section}.
\newline
\textbf{Self-supervised learning.} Self-supervision can be seen as learning by predicting or reconstructing some structured property of the input data itself. For example, in unsupervised models like auto-encoders~\cite{hinton2006reducing,vincent2008extracting}, the reconstruction of the input data is used as a proxy task. In the recent years, many self-supervised approaches have been proposed for learning visual representations. As categorized in~\cite{Larsson_2017_CVPR}, most methods fall into the class of spatial, temporal or colorization-based self-supervision. Spatial self-supervision adopts explaining of a spatial characteristic of an image as a proxy task. Spatial context prediction~\cite{doersch2015unsupervised}, inpainting~\cite{pathak2016context}, solving jigsaw puzzles~\cite{noroozi2016unsupervised} and predicting segmentation maps~\cite{Pathak_2017_CVPR}, all belong to this category. Temporal self-supervising methods model the correlations between video frames as a way to generate supervising signals. Works like~\cite{fernando2016self,misra2016unsupervised,sermanet2017time} model temporal coherence in the frames as supervision while others like~\cite{ranzato2014video,srivastava2015unsupervised} use future frame prediction as a proxy task. Colorization as a proxy task has been experimented with in recent papers of~\cite{Larsson_2017_CVPR} and~\cite{Zhang_2017_CVPR}. In the natural language domain, unsupervised learning of word embeddings by~\cite{mikolov2013distributed} is also an example of self-supervision where prediction of word context is used as a proxy task. Finally, both previously proposed methods for unsupervised visual grounding can also be viewed as self-supervised.~\cite{rohrbach2016grounding} uses reconstruction of one modality using another as a proxy task while~\cite{Xiao_2017_CVPR} exploits regularities in the phrase-structure in addition to a contrastive loss function to create self-supervision. 
Most visual proxy tasks like puzzle solving and inpainting, given their formulation, should ideally learn some high-level representation at the object or scene level. But as noted in the work of~\cite{Bau_2017_CVPR}, self supervised representations typically end up learning low-level features like texture detectors. To the best of our knowledge, we are the first ones to formulate a self-supervised proxy task which is semantic in nature.
\begin{figure*}[t!]
\centering
\includegraphics[width=0.80\textwidth]{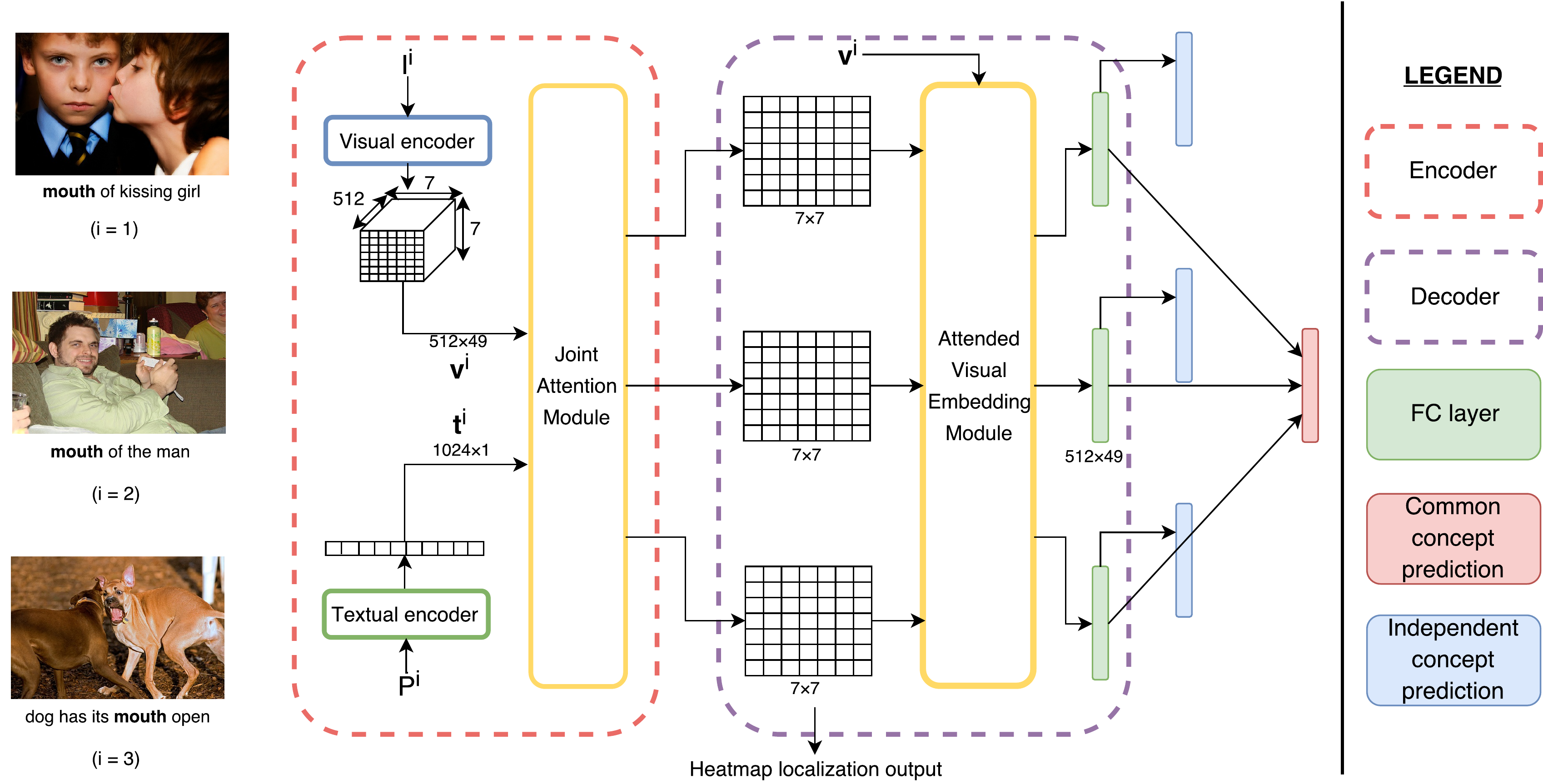}
\caption{An overview of our model for unsupervised visual grounding of phrases. The encoder takes in a set of image-phrase pairs, indexed by $i$, all sharing a common concept. The encoder embeds the image and the phrase to $\mathbf{V}^{i}$ and $\mathbf{t}^{i}$ respectively. These features are used to induce a parametrization for spatial attention. Next, the decoder uses the visual attention map to predict the common concept. In addition, the decoder also predicts the common concept independently for each pair ($i$). For details, see Section~\ref{sec:EDmodel}.}
\label{fig:approach}
\end{figure*}
\section{Grounding through semantic self-supervision}
Unsupervised learning can be interpreted as learning an energy function which assigns lower energy value for data points similar to the training set while assigning high energy value to others. In a self-supervised environment, the role of proxy task is to learn the function that pulls down the energy at the data manifold. With this in mind, we define our proxy task for visual grounding.
\subsection{Proxy task formulation}
Our model is trained for the proxy task of concept-learning. A concept is defined as the entity which is to be grounded in the image. For example, in the phrase \emph{`white \textbf{towel} on the counter'}, the highlighted word \emph{`towel'} is the concept. We observe that in most phrase-image pairs, the localization refers to some concept which explicitly occurs in the phrase as a single word. Though there are some phrases like \emph{`a calm blue \textbf{water body}'} where the concept is multi-word entity, such phrases occur rarely or can be approximated with a single word. We hypothesize that if we induce a parametrization for localization of the phrase and use the localized regions to predict the concept present in an image, the parametrization will converge to the ground truth localization of the phrase. Given this proxy task, we're faced with two main challenges: 1) How do we identify the concept in a phrase? and 2) How do we learn concept representations in an unsupervised setting?

For the first part, we note that identifying a concept which is to be grounded in a phrase, is a problem from the linguistics domain. We can imagine an external system which takes in as input the phrase and returns the concept. Assuming the concept is a single-word entity and exists within the phrase, a naive system can randomly pick a word from the phrase. A more sophisticated sampler can annotate and use the POS tags of the phrase to return the concept. Since most concepts to be localized are nouns, a POS tagger should perform better than random sampling of words. Though better techniques can be employed for concept identification, for the sake of simplicity, we choose the POS tagger to find all nouns in a phrase and randomly select one of them as the concept.

For the second problem, we introduce the notion of a concept batch and learn the concept-prediction task with such batches. A concept batch, as shown in Figure~\ref{fig:approach}, is one training instance for our model, which itself consists of $k$ phrase-image pairs, all containing a common concept. The proxy task is now re-formulated as jointly decoding the common concept using all $k$ localized feature representations in addition to independently decoding the same concept. The intuition behind training with a concept batch is that for decoding the common concept, $k$ phrase-image pairs should encode a localized representation which is invariant to the difference in context across the $k$ pairs. On the other hand, the proxy task of decoding independent concept (for each image in the batch) ensures two things: a) Individual and common representations are consistent b) Model cannot find a shortcut by using only few inputs from the concept batch to decode the common concept.

It is important to note that using a concept batch for learning along with a noun-based concept can be interpreted as generating weak supervision, albeit noisy in nature. Instead of an imperfect concept-identifier, if an oracle could generate a concept which always corresponded to the actual region to be grounded, then this would convert the unsupervised problem to a weakly supervised one. But since the concept identifier might choose the wrong concept or the actual entity to be grounded is not present as a single word in the phrase, the concept-identifier can generate noisy gradients by selecting concepts which might not have anything to do with the localized region. For example, for the phrase \emph{`air-crew boarding the plane'}, the region to be grounded corresponds to the word \emph{`air-crew'} But the sub-optimal concept-identifier might select \emph{`air'} as the concept, which can introduce noise in the weak supervision. Nevertheless, since the same image-phrase pair can be chosen with different sampled concepts during training, it is this random sampling of concepts which ensures that the model doesn't only learn a simple concept-identifier, but also generates information which can help it discriminate between the same concept in different contexts.
\subsection{Encoder-Decoder model}
\label{sec:EDmodel}
We adopt an encoder-decoder architecture for learning to ground as illustrated in Figure \ref{fig:approach}. The encoder uses an attention mechanism similar to~\cite{xu2015show} using the joint features from visual and textual modalities. To maintain fair comparison with previous work, the image features are extracted from the last convolution layer of a VGG16 model~\cite{simonyan2014very} pre-trained on ImageNet. Similarly, the phrase features are extracted from a language model trained on next word prediction on the Google 1 Billion dataset~\cite{chelba2013one} and the MS COCO captions dataset~\cite{lin2014microsoft}. As done in~\cite{Xiao_2017_CVPR}, both the model weights are frozen during training and aren't fine tuned. For the $i^{th}$ index in the concept batch, given visual features from VGG16, $\mathbf{V}^{i} = f_{VGG}(I^{i})$ and textual features from the language model $\mathbf{t}^{i} = f_{LM}(P^{i})$, the attention over visual regions is given by:
\begin{equation}
\small
\mathbf{f}_{attn}^{i}=softmax(\mathbf{f}_{joint}(\mathbf{V^{i}},\mathbf{t^{i}})).
\end{equation}
\begin{equation}\label{fjoint}
\small
\mathbf{f}_{joint}(\mathbf{V}^{i},\mathbf{t}^{i})=\Phi_s(\Phi_r(\Phi_q(\Phi_p([\mathbf{V}^{i},\mathbf{t}^{i}])))),
\end{equation}
where $\mathbf{V}^{i} \in \mathbb{R}^{m \times n}$, $\mathbf{t}^{i} \in \mathbb{R}^{l \times 1}$, $\mathbf{f}_{joint}(\mathbf{V}^{i},\mathbf{t}^{i}) \in \mathbb{R}^{1 \times n}$, $[\mathbf{V}^{i},\mathbf{t}^{i}]$ is an index-wise concatenation operator (over the first dimension) between a matrix $\mathbf{V}^{i}$ and a vector $\mathbf{t}^{i}$ resulting in a matrix of size $((m+l) \times n )$. $\Phi(\cdot)$ corresponds to a hidden layer of a neural network and is defined as:
\begin{equation}
\small
\Phi_p(\mathbf{X})=ReLU(\mathbf{W}_p \mathbf{X} + \mathbf{b}_p),
\end{equation}
where $ReLU(x)=max(x,0)$, $\mathbf{W}_p \in \mathbb{R}^{p \times d}$, $\mathbf{b}_p \in \mathbb{R}^{p \times 1}$ and $\mathbf{X} \in \mathbb{R}^{d \times n}$. Here $n$ is the number of regions over which attention is defined and $d$ is the dimensionality of each region with respective to $\mathbf{X}$. 

Thus, we use a 4 layered non linear perceptron to calculate attention for each of the $n$ regions \footnote{Since we compute attention over VGG feature maps, $n$ = $7\times7$ }. In contrast to~\cite{rohrbach2016grounding}, we compute attention over the spatial regions of the last feature maps from VGG16 instead of computing it over bounding boxes. The four $\Phi(\cdot)$ layers gradually decrease the dimensionality of the concatenated joint features from $(m+l) \to p \to q \to r \to s$ where $s=1$. It is important to note that the attention module is shared across all $\mathbf{V}^{i}$ and $\mathbf{t}^{i}$. Thus the encoder is common for all pairs in the concept batch. Next, we describe a decoding mechanism to predict the common and independent concept. 

Given the attention weights $\mathbf{f}_{attn}^{i} \in \mathbb{R}^{1 \times n}$, the visual attention for common concept prediction ($\mathbf{f}_{vac}$) is computed by taking the weighted sum with the original visual features.
\begin{equation}
\small
\mathbf{f}_{vac}=\sum_{i=1}^{k} \mathbf{f}_{attn}^{i} \mathbf{V}^{i}
\end{equation}
We find that aggregating the visual attention across regions, which is commonly done in the past attention literature degrades performance for our task. Therefore we retain the spatial information and only aggregate the features across the concept batch.

Similarly, the visual attention for independent concept prediction, $\mathbf{f}_{vai}^{i}$ is given by the element-wise product of the attention weights and visual features.
\begin{equation}
\small
\mathbf{f}_{vai}^{i}=\mathbf{f}_{attn}^{i} \mathbf{V}^{i}
\end{equation}
Finally, both the attended features are flattened and separately connected to a fully connected layer, leading to a softmax over the concepts. In practice, we also down-sample the dimensionality of $\mathbf{f}_{vai}^{i}$ using $1*1$ convolutions before we aggregate and flatten the features.
\begin{equation}
\small
\mathbf{y}_{common}=softmax(\mathbf{W}_{vac} \mathbf{f}_{vac} + \mathbf{b}_{vac}).
\end{equation}
\begin{equation}
\small
\mathbf{y}_{independent}^{i}=softmax(\mathbf{W}_{vai} \mathbf{f}_{vai}^{i} + \mathbf{b}_{vai}),
\end{equation}
where $\mathbf{y}_{common}$ is the network prediction for the common concept and $\mathbf{y}_{independent}^{i}$ is the independent concept prediction for the $i^{th}$ index in the concept batch.

\subsection{Surrogate loss}
Our surrogate loss consists of two different terms, one corresponding to the common concept prediction and the other for independent concept prediction. Since we decode the visually attended features to a softmax over the concept vocabulary, we use the cross-entropy loss to train our model. Given the target common concept vector $\mathbf{y}_{t}$ for a concept batch of size $k$, the proxy objective function is:
\begin{equation}\label{loss_total}
\small
L_{total} = L(\mathbf{y}_{common}, \mathbf{y}_{t}) + \dfrac{1}{k}\sum_{i=1}^{k} L(\mathbf{y}_{independent}^{i}, \mathbf{y}_{t})
\end{equation}
where $L(\cdot)$ is the standard cross-entropy loss.
\begin{table}[t]
  \centering
 \scalebox{0.75}{ \begin{tabular}{lSSS}
    \toprule
    \textbf{Dataset Statistics} & \multicolumn{3}{c}{\textbf{Value}}  \\
    & {Visual Genome} & {ReferIt} & {Flickr30k} \\
    \midrule
    No of phrases per image & 50.0 & 5.0 & 8.7\\
    No of objects per image & 35.0 & {-} & 8.9\\
    Word count per phrase & 5.0 & 3.4 & 2.3\\
    Noun count per phrase & 2.2 & 1.8 & 1.2\\
    \bottomrule
  \end{tabular}}
  \vspace{0.2em}
  \caption{Phrase-region related statistics for datasets used in evaluation. The numbers reflect the relative complexity of these datasets.}
\label{table:dataset_stats}
\end{table}
\section{Experimental setup}
In this section, we elaborate upon the implementation details, employed datasets, evaluation metric and baselines.
\subsection{Implementation details}
An ImageNet pre-trained VGG16 and a Google 1 Billion trained language model are used for encoding the image and the phrase respectively. Both the visual and textual feature extractors are fixed during training. Before the attention module, both the features are normalized using a batch-normalization layer~\cite{ioffe2015batch}. The concept vocabulary used for training with the softmax based loss is taken from the most frequently occurring nouns. Since the frequency distribution follows the Zipf's Law, around 95\% of the phrases are accounted for by the top 2000 concepts, which is used as the softmax size. Another implication of this distribution is that random sampling of instances skews the number of times a concept is seen during training. Therefore instead of creating mini batches by randomly sampling instances, we randomly sample a concept from the concept vocabulary and using this, we randomly sample $k$ phrase-image pairs where this word occurs, thus creating our concept batch. In the encoder, the values of $p, q, r, s$ from Equation ~\ref{fjoint} are taken as $512, 128, 32, 1$ respectively. We train our models using the Adam optimizer~\cite{kingma2014adam} with a batch size of $16$. 
\begin{table*}[t]
  \centering
  \scalebox{0.75}{\begin{tabular}{lSSSS}
    \toprule
    \textbf{Method} & \multicolumn{4}{c}{\textbf{Accuracy}}  \\
    & {Visual Genome} & {ReferIt (mask)} & {ReferIt (bbox)} & {Flickr30k} \\
    \midrule
    Random baseline & 11.15 & 16.48 & 24.30 & 27.24\\
    Center baseline & 20.55 & 17.04 & 30.40 & 49.20\\
    VGG baseline & 18.04 & 15.64 & 29.88 & 35.37\\
    Fang \emph{et al.}~\cite{fang2015captions} & 14.03 & 23.93 & 		33.52 & 29.03\\
    Zhang \emph{et al.}~\cite{zhang2016top} & 19.31 & 21.94 & 			31.97 & 42.40\\
    Ramanishka \emph{et al.}~\cite{Ramanishka2017cvpr} & {-} & {-} & {-} & {\textbf{50.10}}\\
    Xiao \emph{et al.}~\cite{Xiao_2017_CVPR} & 24.40 & {-} & {-} & {-}\\
    Semantic self-supervision (Ours) & {\textbf{30.03}} & {\textbf{29.72}} & {\textbf{39.98}} & 49.10\\
    \bottomrule
  \end{tabular}}
  \vspace{0.2em}
  \caption{Phrase grounding evaluation on 3 datasets using the pointing game metric. See Section~\ref{sec:result_section} for mask vs bbox explanation for ReferIt.}
\label{table:results}
\end{table*}
\subsection{Evaluation}
\textbf{Dataset.}
We test our method on the Visual Genome~\cite{krishna2017visual}, the ReferItGame~\cite{KazemzadehOrdonezMattenBergEMNLP14} and the Flickr30k Entities~\cite{plummer2015flickr30k} datasets and there exist few important qualitative and quantitative differences between them.
Visual Genome has a longer average description length than ReferIt which is in turn higher than Flickr30k. As mentioned in~\cite{krishna2017visual}, Visual Genome and ReferIt include description for regions which are less salient and hence harder to localize unlike Flickr30k. On the other hand, Flickr30k provides multiple bounding box annotations for different description instances within an image. So for a phrase, \emph{`trees behind the lake'}, all instances of \emph{`trees'} are annotated separately in Flickr30k whereas Visual Genome only annotates a single instance of the description. Unlike the others, ReferIt usually only refers to a specific object instance. Table~\ref{table:dataset_stats} shows some important dataset statistics which hint towards the complexity of the datasets. For example, notice that in Flickr30k, the average phrase length is just $2.3$ words and average noun count is $1.2$ which would mean that the region to be localized in most cases is directly present as a single word, thus changing the problem to an almost weakly supervised setting. To ensure fair comparison with the previous work of~\cite{Xiao_2017_CVPR}, we use the images from the validation set of MS-COCO which have region annotations in the Visual Genome dataset as our test set. We use remaining images of the Visual Genome dataset for training. For ReferIt and Flickr30k, we use the test sets for evaluation.
\newline
\textbf{Evaluation metric.}
Since our model generates localization in the form of a heatmap, we evaluate our model with the pointing game metric~\cite{zhang2016top}, similar to the previous work of~\cite{Xiao_2017_CVPR, Ramanishka2017cvpr}. Pointing game measures how accurate the most confident region in the predicted heatmap is with respect to the ground truth bounding box. For a given input image-phrase pair, the predicted localization heatmap is considered a \emph{Hit} if the pixel with the maximum value lies within the bounding box, else it's considered a \emph{Miss}. Therefore it purely evaluates the spatial accuracy of the heatmap instead of measuring the extent of localization. The pointing game accuracy is defined as the fraction of correct localizations out of the total testing instances, i.e.\ $\tfrac{\#Hit}{\#Hit + \#Miss}$. For an image of size $224\times224$, the $7\times7$ attention map is projected back using a stride of $224/7$. Thus each of the $49$ grids correspond to a $32\times32$ region in the original image space and the center point of the highest activated grid is chosen as the maximum value for the pointing game.
For the purpose of visualization, we use bilinear interpolation of the attention weights to generate image-sized heatmap.
\newline
\textbf{Baselines.}
We compare the performance of our approach with multiple baselines and previous methods. The first is a random baseline which mimics the attention-based localization of our setup, but chooses the region randomly. This baseline randomly chooses one out of the $49$ regions. The second baseline is taken from~\cite{Ramanishka2017cvpr,zhang2016top} where the center point of the image is taken as the max for the pointing game. Note that this baseline can produce skewed results in datasets where the phrase to be localized has a center-bias, which is what we observe with Flickr30k (as previously noted in~\cite{Ramanishka2017cvpr}). We also use a visual-only baseline which selects the maximum pixel value for the pointing game on the basis of pre-trained visual features. For this, we use the feature maps from the last convolution layer of an ImageNet pre-trained VGG16 and average the channel activations to get a $7\times7$ map. We then choose the maximum activated grid for the pointing game.
Apart from these three baselines, we also compare against weakly supervised works of~\cite{fang2015captions} and~\cite{zhang2016top} who use an MIL based approach and an excitation backprop scheme respectively for localizing single-word labels. As done in~\cite{zhang2016top,Ramanishka2017cvpr}, we average the heatmaps generated for tokens present in their dictionary for obtaining the final heatmap. Finally, we also compare against the more recent unsupervised works of~\cite{Xiao_2017_CVPR,Ramanishka2017cvpr}.
\section{Results}
\label{sec:result_section}
We report the comparison of our method with the baselines and previous methods in this section. Table~\ref{table:results} summarizes the performance of our best model on the three datasets. To highlight our generalization ability, we train the proposed model on Visual Genome since it's the largest, more complex dataset out of the three and directly evaluate on the test set of all three datasets without fine tuning.
\newline
\textbf{Visual Genome.} The random baseline yields the least performance as expected. Surprisingly, the VGG16 baseline fares decently well given that it does not take any phrase-related information into account. We believe this is due to the phrases often referring to some object in the image which the VGG16 features are already trained for recognizing. Our model outperforms all the baselines and improves upon the previous state-of-art work by~\cite{Xiao_2017_CVPR} by $5.63\%$.
\newline
\textbf{ReferItGame.} 
~\cite{hu2016natural} provide segmentation mask for each phrase-region pair and use them to obtain a bounding box (bbox) which envelopes the mask completely. They then use this for their evaluation on ReferIt. Though we provide evaluation for both bbox and mask settings, we believe that the mask based annotations are more precise and accurate for measuring localization performance. Since both Visual Genome and ReferIt contain phrases which: a) refer to very specific regions like \emph{`red car on corner'} and b) refer to non-salient objects like \emph{`white crack in the sidewalk'}, both datasets have low performance with baselines like center and VGG. Our model outperforms all baselines on ReferIt too, improving upon the MIL based approach by $5.79\%$.
\newline
\textbf{Flickr30k Entities.}
Flickr30k dataset has higher performance across methods as compared to the other two datasets due to the two points mentioned in the previous subsection along with the fact that Flickr30k annotates all bboxes referring to a phrase as opposed to the other datasets which only have a one-to-one phrase-bbox mapping for an image. Our model outperforms most baselines and is just $1\%$ less than the state-of-art work of~\cite{Ramanishka2017cvpr}.
\section{Analysis of the approach}
In this section, we examine the effects of changing the hyperparameter $k$ (concept batch size), the significance of the two surrogate losses and the effect of the concepts with which our model is trained, followed by some qualitative outputs of our model. All the analysis in the following sections is done on the Visual Genome dataset.
\begin{table}[t]
  \centering
  \scalebox{0.80}{\begin{tabular}{lSSSS}
    \toprule
    \textbf{Loss Type} & \multicolumn{4}{c}{\textbf{Concept Batch Size ($\textbf{k}$)}}  \\
    & {$k=3$} & {$k=5$} & {$k=7$} & {$k=9$}\\
    \midrule
    Independent concept only & 27.15 & 27.27 & 28.01 & 28.05 \\
	Common concept only & 27.52 & 28.94 & 29.18 & 27.90 \\
	Independent and common concept & 28.25 & 28.91 & 29.89 & 30.03 \\
    \bottomrule
  \end{tabular}}
  \vspace{0.2em}
  \caption{Analysis of different surrogate losses while varying the concept batch size.}
\label{table:ablations}
\end{table}
\subsection{Concept batch size and surrogate loss}
We perform ablative studies on the two loss terms and the concept batch size $k$ and observe certain patterns. For the discussion in this section, we use the shorthand $IC$ (independent concept only), $CC$ (common concept only) and $ICC$ (independent and common concept) for the three loss types from Table~\ref{table:ablations}. We train our model with the $IC$ and $CC$ loss separately, keeping everything else in the pipeline fixed. This means that the sampling procedure for a concept batch remains intact for all loss types, even when there is no common concept being decoded. For all three settings, we vary the concept batch size $k$ and observe some interesting trends. As shown in Table~\ref{table:ablations}, for a fixed loss type, the performance increases as we increase $k$, the CC loss being the exception to this trend. The performance for $CC$ loss increases up to $k=7$, but goes down with $k=9$. This points to a common problem with self-supervised techniques where the model finds a shortcut to reduce the loss value without improving on the performance. With only the common concept loss, the network can learn a majority voting mechanism such that not all $k$ concept representations need to be consistent with the common concept. Thus, the network can easily optimize the proxy objective, but is not forced to learn a robust grounding for all instances in the concept batch. This is corroborated with the fact that during training, we also observe a faster convergence of $CC$ loss for $k=9$ than the lower concept batch sizes. This empirically highlights the importance of the $IC$ loss term. It also highlights the usefulness of the concept batch formulation since it improves performance in general.
For a fixed $k$, we also observe an expected pattern. $IC$ loss usually achieves the least performance out of the three, with $CC$ loss coming in next. The best performance is obtained with both the losses together. Lastly, we point out the slight increase in performance for $IC$ loss as we increase $k$. We believe that even though this loss type does not jointly decode the common concept explicitly, within a training mini-batch, the gradients which are backpropagated are from the average loss of a concept batch having a common concept (see second term of Equation ~\ref{loss_total}). Thus training with $IC$ loss is not exactly the same as training with a batch of independent image-phrase pairs and simply decoding a concept from the phrase.
\begin{figure}[t]
        \centering
        \scriptsize
        \begin{tabular}[b]{c c}
        \hspace{-1em} \includegraphics[width=0.47\linewidth]{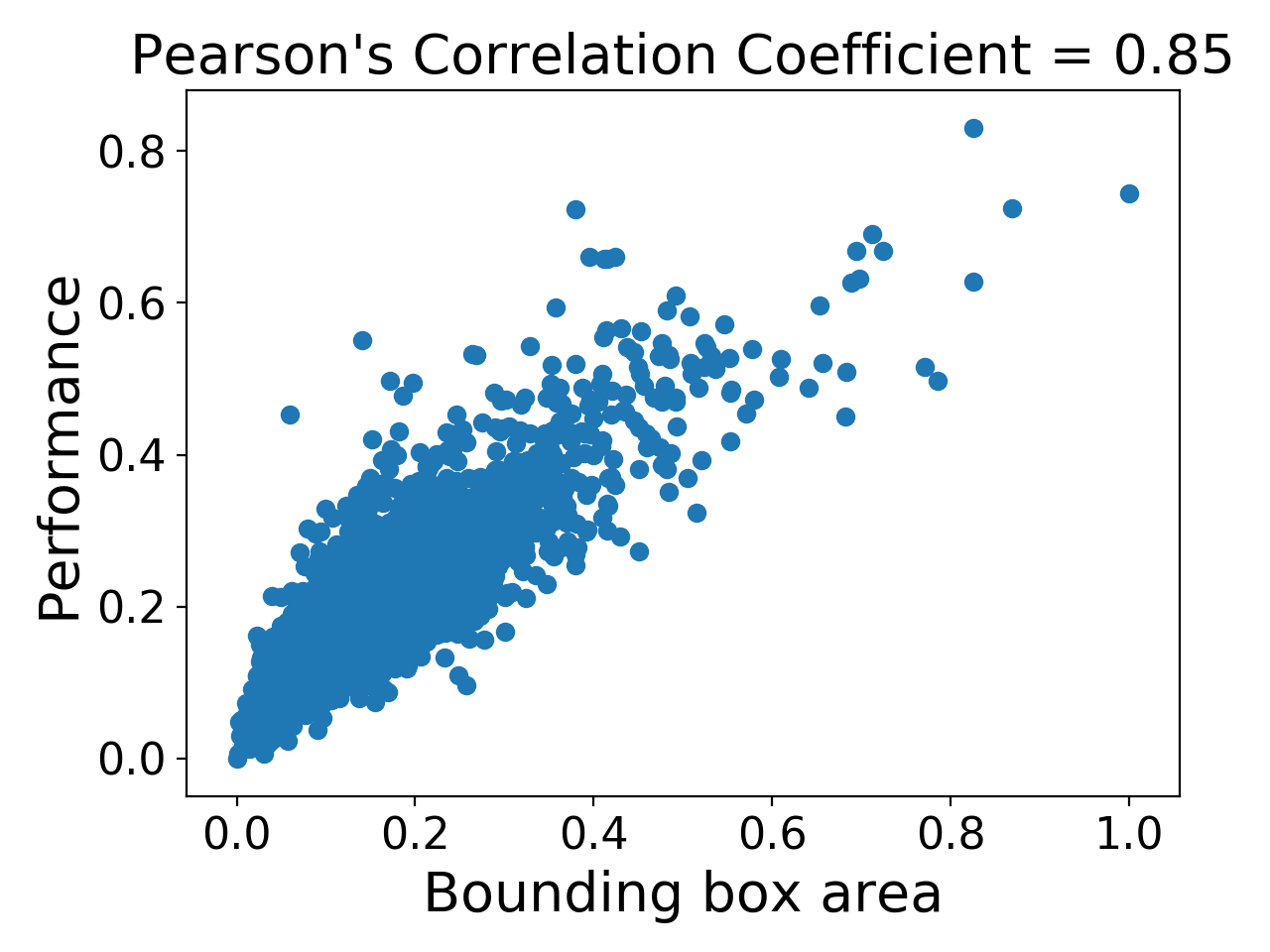}& 
        \includegraphics[width=0.47\linewidth]{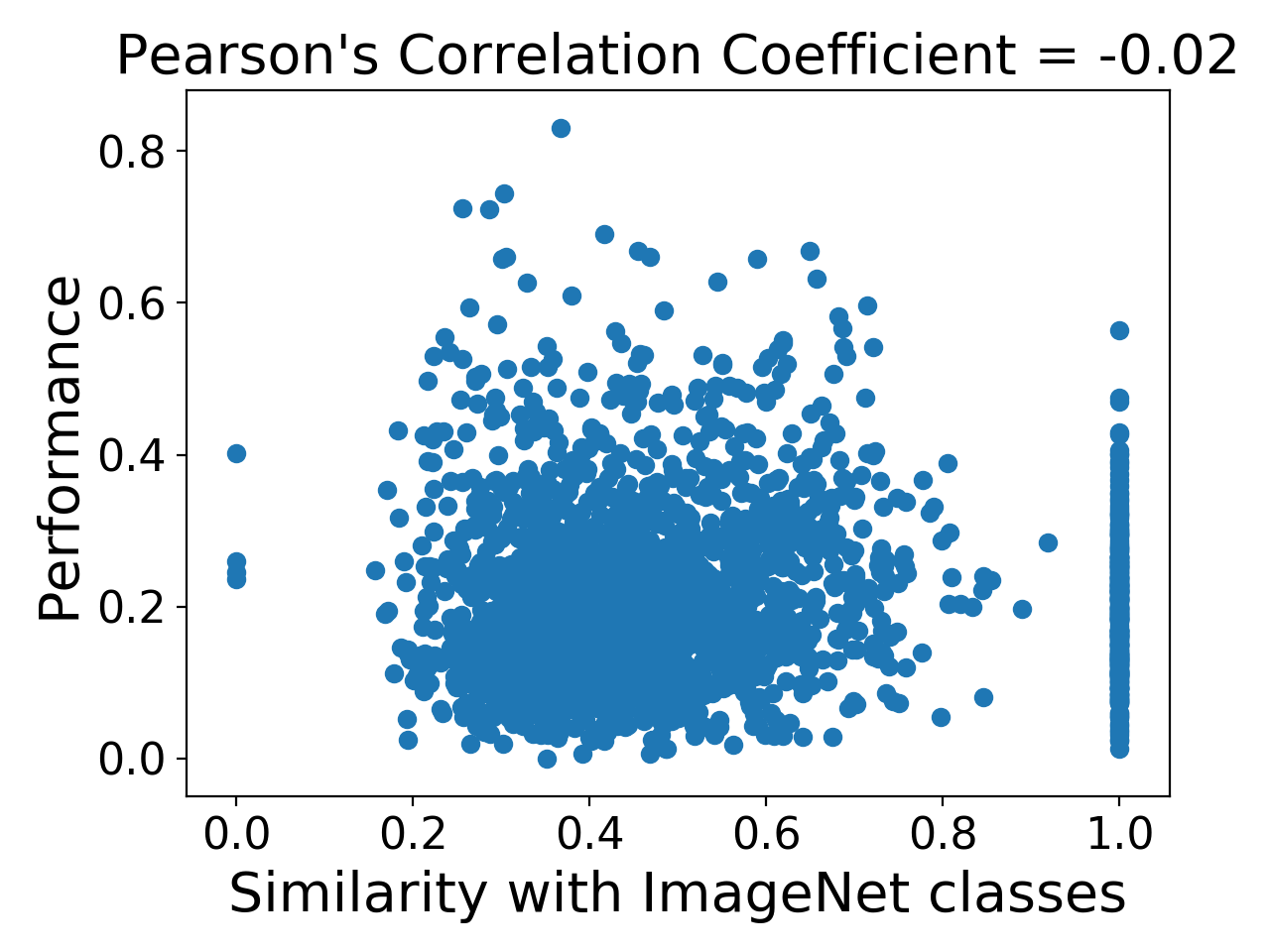} \\
        \end{tabular}
        \caption{Variation of performance with respect to bounding box area and similarity of concept with ImageNet classes.}
        \label{fig:scatter_plots}
\end{figure}
\subsection{Performance variation across concepts}
To better understand the variation in performance across the chosen concepts, we also compute the performance across each of the $2000$ concept classes. We observe a trend in the performance with concepts like \emph{`suitcase'}, \emph{`airplanes'} and \emph{`breakfast'} getting close to $70\%$ accuracy while concepts like \emph{`screw'}, \emph{`socket'} and \emph{`doorknob'} getting less than $5\%$. We investigate two possible causes for this variability. The first is the average bounding box size associated with each of these concepts. The second is the existing knowledge of concept labels present in the ImageNet classes which our model obtains through the VGG16 based visual encoder.
Figure \ref{fig:scatter_plots} (left) shows the variation of performance with respect to the average bounding box area for each concept. We observe a strong positive correlation between the two variables, explaining the lower performance for concepts with small sizes. For computing the correlation of concept performance with the knowledge from ImageNet classes, we use a trained word2vec model~\cite{mikolov2013distributed} and compute the maximum similarity of a particular concept across all the ImageNet classes. We plot this in Figure \ref{fig:scatter_plots} (right) which illustrates no noticeable correlation between the two variables. This further strengthens the case for our approach since we observe that our concept performance isn't biased towards the labels present in ImageNet. 
\begin{figure*}[t]
\centering
\includegraphics[width=0.76\textwidth]{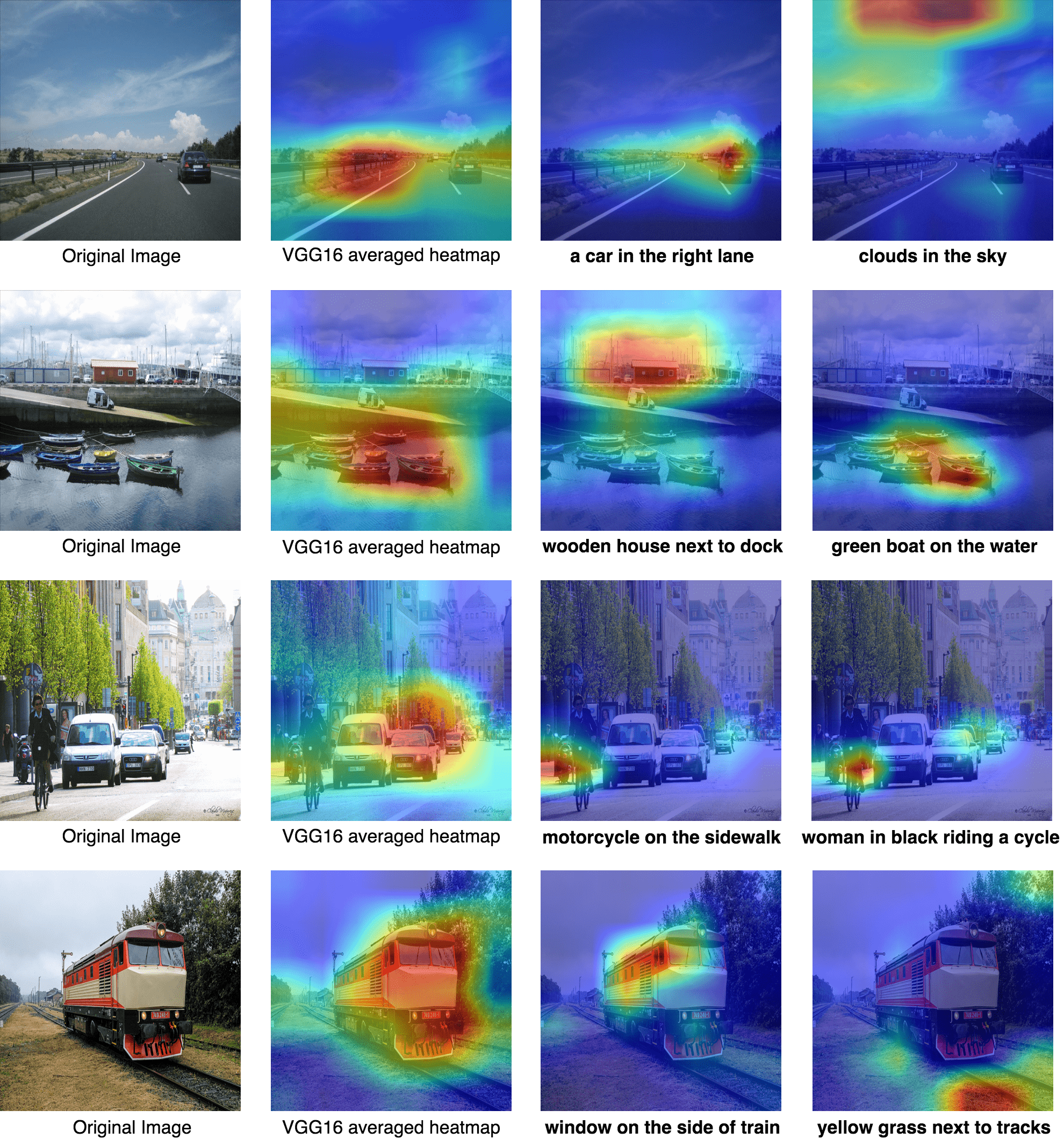}
\caption{Qualitative results of our approach with different image and phrase pairs as input. More results and visual error analysis shown in supplementary material.}
\label{fig:qualitative}
\end{figure*}
\subsection{Improvement over a noun-based concept detector}
We also conduct a small experiment to verify that the model isn't simply working as a noun-based concept detector instead of modeling the complete phrase. For this, we replace the phrase with a single noun randomly sampled from the phrase, as the input to the textual encoder. We note a $4.7\%$ drop in performance on Visual Genome for $k=5$. Since training of the original model enforces only concept-level discrimination, it's interesting to see that presence of complete phrases is useful for performance. This shows that our model is much more than a word-level concept-detector and utilizes the full phrase for grounding.
\subsection{Qualitative analysis}
In this section we show some of our qualitative results on the Visual Genome dataset. Figure \ref{fig:qualitative} shows the localization heatmap we obtain from our attention weights. It also shows the VGG16 activation heatmaps obtained by averaging over the channels. We find that our model doesn't simply generate a phrase-independent saliency map, but focuses even on non-salient regions if the phrase refers to it. This is also evident through comparisons with the VGG maps. We provide many more examples and typical failure cases in the supplementary.
\section{Conclusion}
We propose a novel approach for visual grounding of phrases through a self-supervising proxy task formulation. Our qualitative and quantitative results point to the fact that many semantic regularities exist in the data which can be exploited to learn unsupervised representations for a variety of tasks. Thorough analysis of our model reveals interesting insights which may be useful for future research efforts in the area. Using our approach, we achieve state-of-art performance on multiple datasets. Finally, we note that as the complexity of these visual tasks increase, the role of language can become pivotal in learning richer representations, the task of unsupervised grounding being one of the case in point.
\clearpage
{\small
\bibliographystyle{ieee}
\bibliography{egbib}

\begin{thebibliography}{10}\itemsep=-1pt

\bibitem{Bau_2017_CVPR}
D.~Bau, B.~Zhou, A.~Khosla, A.~Oliva, and A.~Torralba.
\newblock Network dissection: Quantifying interpretability of deep visual
  representations.
\newblock In {\em CVPR}, 2017.

\bibitem{chelba2013one}
C.~Chelba, T.~Mikolov, M.~Schuster, Q.~Ge, T.~Brants, P.~Koehn, and
  T.~Robinson.
\newblock One billion word benchmark for measuring progress in statistical
  language modeling.
\newblock {\em arXiv preprint}, 2013.

\bibitem{Chen_2017_ICCV}
K.~Chen, R.~Kovvuri, and R.~Nevatia.
\newblock Query-guided regression network with context policy for phrase
  grounding.
\newblock In {\em ICCV}, 2017.

\bibitem{doersch2015unsupervised}
C.~Doersch, A.~Gupta, and A.~A. Efros.
\newblock Unsupervised visual representation learning by context prediction.
\newblock In {\em ICCV}, 2015.

\bibitem{fang2015captions}
H.~Fang, S.~Gupta, F.~Iandola, R.~K. Srivastava, L.~Deng, P.~Doll{\'a}r,
  J.~Gao, X.~He, M.~Mitchell, J.~C. Platt, et~al.
\newblock From captions to visual concepts and back.
\newblock In {\em CVPR}, 2015.

\bibitem{fernando2016self}
B.~Fernando, H.~Bilen, E.~Gavves, and S.~Gould.
\newblock Self-supervised video representation learning with odd-one-out
  networks.
\newblock {\em arXiv preprint}, 2016.

\bibitem{frome2013devise}
A.~Frome, G.~S. Corrado, J.~Shlens, S.~Bengio, J.~Dean, T.~Mikolov, et~al.
\newblock Devise: A deep visual-semantic embedding model.
\newblock In {\em NIPS}, 2013.

\bibitem{fukui2016multimodal}
A.~Fukui, D.~H. Park, D.~Yang, A.~Rohrbach, T.~Darrell, and M.~Rohrbach.
\newblock Multimodal compact bilinear pooling for visual question answering and
  visual grounding.
\newblock {\em arXiv preprint}, 2016.

\bibitem{hinton2006reducing}
G.~E. Hinton and R.~R. Salakhutdinov.
\newblock Reducing the dimensionality of data with neural networks.
\newblock {\em Science}, 2006.

\bibitem{hu2016natural}
R.~Hu, H.~Xu, M.~Rohrbach, J.~Feng, K.~Saenko, and T.~Darrell.
\newblock Natural language object retrieval.
\newblock In {\em CVPR}, 2016.

\bibitem{ioffe2015batch}
S.~Ioffe and C.~Szegedy.
\newblock Batch normalization: Accelerating deep network training by reducing
  internal covariate shift.
\newblock In {\em ICML}, 2015.

\bibitem{KazemzadehOrdonezMattenBergEMNLP14}
S.~Kazemzadeh, V.~Ordonez, M.~Matten, and T.~L. Berg.
\newblock Referit game: Referring to objects in photographs of natural scenes.
\newblock In {\em EMNLP}, 2014.

\bibitem{kingma2014adam}
D.~Kingma and J.~Ba.
\newblock Adam: A method for stochastic optimization.
\newblock {\em arXiv preprint}, 2014.

\bibitem{kiros2014unifying}
R.~Kiros, R.~Salakhutdinov, and R.~S. Zemel.
\newblock Unifying visual-semantic embeddings with multimodal neural language
  models.
\newblock {\em arXiv preprint}, 2014.

\bibitem{krishna2017visual}
R.~Krishna, Y.~Zhu, O.~Groth, J.~Johnson, et~al.
\newblock Visual genome: Connecting language and vision using crowdsourced
  dense image annotations.
\newblock {\em IJCV}, 2017.

\bibitem{Larsson_2017_CVPR}
G.~Larsson, M.~Maire, and G.~Shakhnarovich.
\newblock Colorization as a proxy task for visual understanding.
\newblock In {\em CVPR}, 2017.

\bibitem{lin2014microsoft}
T.-Y. Lin, M.~Maire, S.~Belongie, et~al.
\newblock Microsoft coco: Common objects in context.
\newblock In {\em ECCV}, 2014.

\bibitem{mikolov2013distributed}
T.~Mikolov, I.~Sutskever, K.~Chen, G.~S. Corrado, and J.~Dean.
\newblock Distributed representations of words and phrases and their
  compositionality.
\newblock In {\em NIPS}, 2013.

\bibitem{misra2016unsupervised}
I.~Misra, C.~L. Zitnick, and M.~Hebert.
\newblock {Shuffle and Learn: Unsupervised Learning using Temporal Order
  Verification}.
\newblock In {\em ECCV}, 2016.

\bibitem{noroozi2016unsupervised}
M.~Noroozi and P.~Favaro.
\newblock Unsupervised learning of visual representations by solving jigsaw
  puzzles.
\newblock In {\em ECCV}, 2016.

\bibitem{Pathak_2017_CVPR}
D.~Pathak, R.~Girshick, P.~Dollar, T.~Darrell, and B.~Hariharan.
\newblock Learning features by watching objects move.
\newblock In {\em CVPR}, 2017.

\bibitem{pathak2016context}
D.~Pathak, P.~Krahenbuhl, J.~Donahue, T.~Darrell, and A.~A. Efros.
\newblock Context encoders: Feature learning by inpainting.
\newblock In {\em CVPR}, 2016.

\bibitem{plummer2017phrase}
B.~A. Plummer, A.~Mallya, C.~M. Cervantes, J.~Hockenmaier, and S.~Lazebnik.
\newblock Phrase localization and visual relationship detection with
  comprehensive image-language cues.
\newblock In {\em CVPR}, 2017.

\bibitem{plummer2015flickr30k}
B.~A. Plummer, L.~Wang, C.~M. Cervantes, J.~C. Caicedo, J.~Hockenmaier, and
  S.~Lazebnik.
\newblock Flickr30k entities: Collecting region-to-phrase correspondences for
  richer image-to-sentence models.
\newblock In {\em ICCV}, 2015.

\bibitem{Ramanishka2017cvpr}
V.~Ramanishka, A.~Das, J.~Zhang, and K.~Saenko.
\newblock Top-down visual saliency guided by captions.
\newblock In {\em CVPR}, 2017.

\bibitem{ranzato2014video}
M.~Ranzato, A.~Szlam, J.~Bruna, M.~Mathieu, R.~Collobert, and S.~Chopra.
\newblock Video (language) modeling: a baseline for generative models of
  natural videos.
\newblock {\em arXiv preprint}, 2014.

\bibitem{Ren_etal_BMVC_17}
Z.~Ren, H.~Jin, Z.~Lin, C.~Fang, and A.~Yuille.
\newblock Multiple instance visual-semantic embedding.
\newblock In {\em BMVC}, 2017.

\bibitem{rohrbach2016grounding}
A.~Rohrbach, M.~Rohrbach, R.~Hu, T.~Darrell, and B.~Schiele.
\newblock Grounding of textual phrases in images by reconstruction.
\newblock In {\em ECCV}, 2016.

\bibitem{sermanet2017time}
P.~Sermanet, C.~Lynch, J.~Hsu, and S.~Levine.
\newblock Time-contrastive networks: Self-supervised learning from multi-view
  observation.
\newblock {\em arXiv preprint}, 2017.

\bibitem{simonyan2014very}
K.~Simonyan and A.~Zisserman.
\newblock Very deep convolutional networks for large-scale image recognition.
\newblock {\em arXiv preprint}, 2014.

\bibitem{srivastava2015unsupervised}
N.~Srivastava, E.~Mansimov, and R.~Salakhudinov.
\newblock Unsupervised learning of video representations using lstms.
\newblock In {\em ICML}, 2015.

\bibitem{vincent2008extracting}
P.~Vincent, H.~Larochelle, Y.~Bengio, and P.-A. Manzagol.
\newblock Extracting and composing robust features with denoising autoencoders.
\newblock In {\em ICML}, 2008.

\bibitem{wang2016learning}
L.~Wang, Y.~Li, and S.~Lazebnik.
\newblock Learning deep structure-preserving image-text embeddings.
\newblock In {\em CVPR}, 2016.

\bibitem{Xiao_2017_CVPR}
F.~Xiao, L.~Sigal, and Y.~Jae~Lee.
\newblock Weakly-supervised visual grounding of phrases with linguistic
  structures.
\newblock In {\em CVPR}, 2017.

\bibitem{xu2015show}
K.~Xu, J.~Ba, R.~Kiros, K.~Cho, A.~Courville, R.~Salakhudinov, R.~Zemel, and
  Y.~Bengio.
\newblock Show, attend and tell: Neural image caption generation with visual
  attention.
\newblock In {\em ICML}, 2015.

\bibitem{zhang2016top}
J.~Zhang, Z.~Lin, J.~Brandt, X.~Shen, and S.~Sclaroff.
\newblock Top-down neural attention by excitation backprop.
\newblock In {\em ECCV}, 2016.

\bibitem{Zhang_2017_CVPR}
R.~Zhang, P.~Isola, and A.~A. Efros.
\newblock Split-brain autoencoders: Unsupervised learning by cross-channel
  prediction.
\newblock In {\em CVPR}, 2017.

\end{thebibliography}
}
\clearpage

\section*{\centering Supplementary Material}

\section{Introduction}
\vspace{-0.5em}
In the supplementary material, we qualitatively highlight the following:
\vspace{-0.5em}
\begin{itemize}
  \setlength\itemsep{0.05em}
  \item How our generated heatmap differs from the VGG16 features which are used as visual input to our model.
  \item How the alignment of the chosen ground truth concept, the predicted concept and the actual entity to be grounded affects the quality of the phrase grounding.
\end{itemize}
\vspace{-1.5em}
\section{Comparison with VGG16 feature maps}
\begin{figure*}[thpb]
\centering
\vspace{1em}
\includegraphics[width=\textwidth]{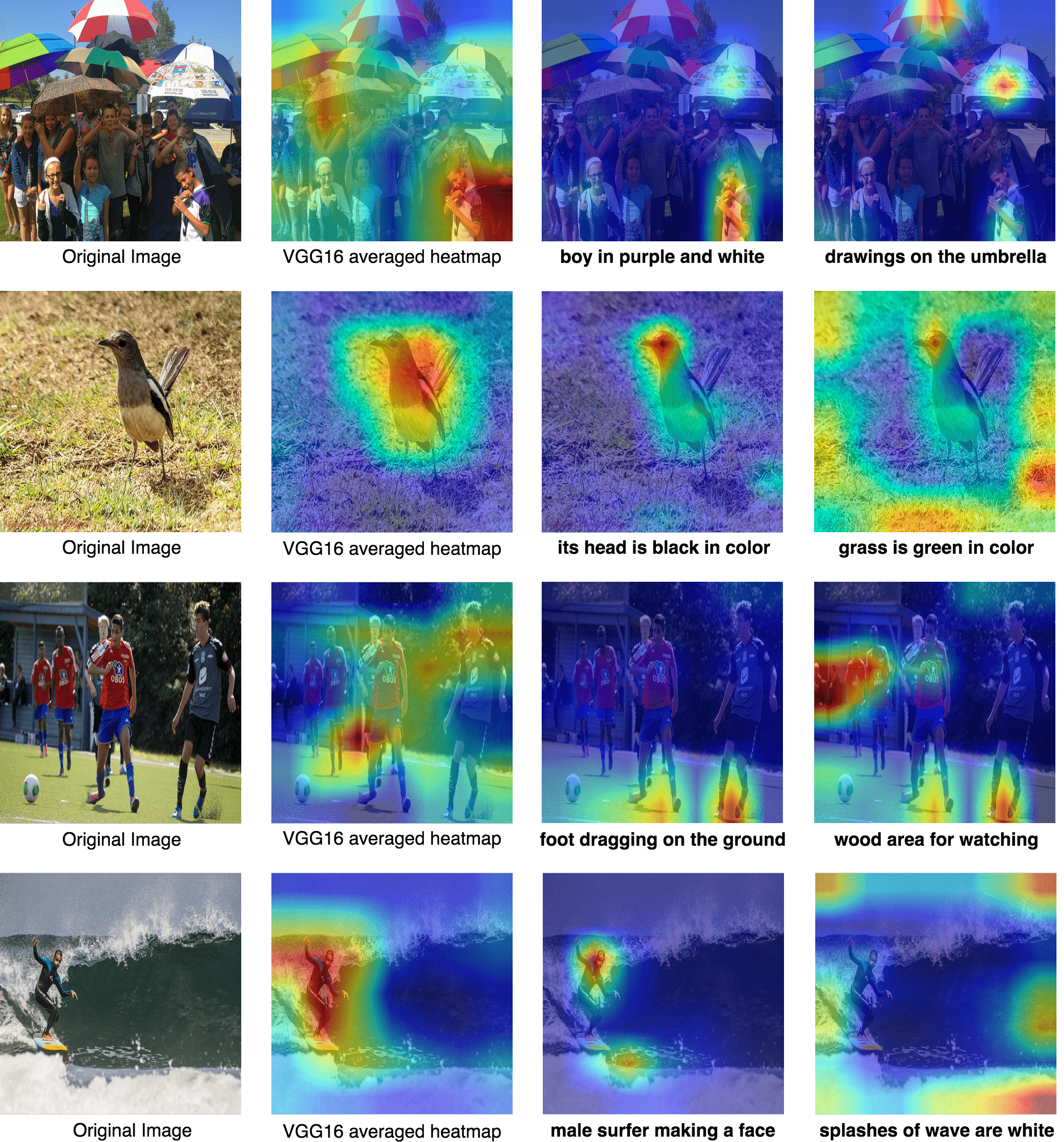}
\caption{Comparison of VGG16 feature maps with our generated attention maps.}
\label{fig:fig1}
\end{figure*}

\vspace{-0.5em}
Our model uses a pre-trained VGG16 model to extract feature maps for our visual encoder. In this section we show that even though the visual features are fixed during training, our model learns attention maps which are spatially distinct from the VGG16 feature maps used as input. We use the channel averaged VGG16 baseline model for visualizing the VGG16 heatmap. Figure~\ref{fig:fig1} shows the comparison between this and our predictions. As evident from the examples, our method produces attention maps which can localize regions which were weak or even non-existent in the activations of the input maps. This shows that our model doesn't simply amplify the activations present in VGG16 channels but learns a phrase dependent attention map.
\vspace{-0.5em}
\section{Alignment of the selected, predicted and true concept}
\label{sec:alignment}
\begin{figure*}[thpb]
\centering
\vspace{1em}
\includegraphics[width=0.9\textwidth]{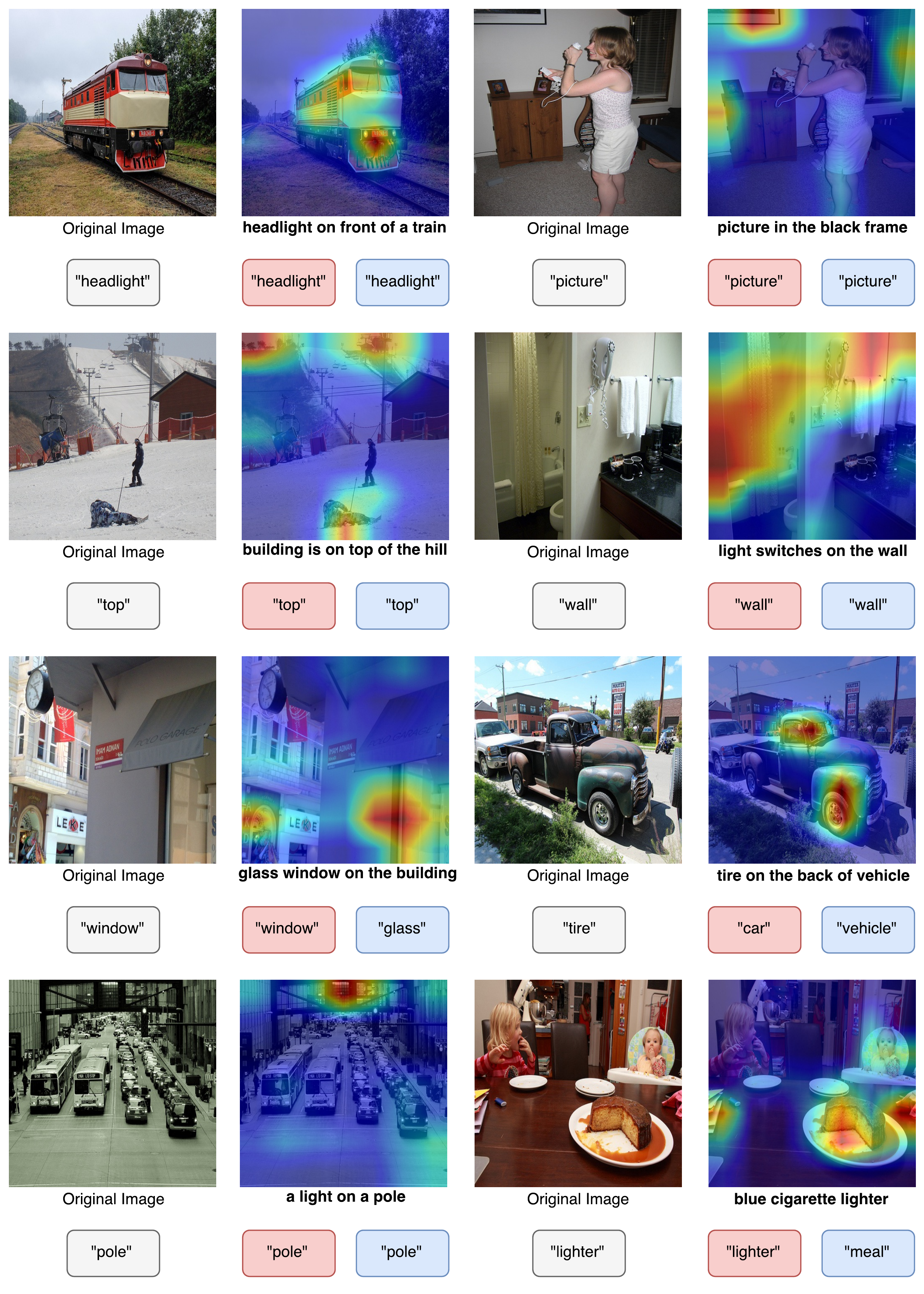}
\caption{The figure shows how the quality of output heatmap changes with the alignment of the selected concept, predicted concept and the real entity to be grounded. For some sampled concept batch, the \textbf{\textcolor{selected_concept_color}{gray box}} refers to the chosen common concept,  the \textbf{\textcolor{common_concept_color}{red box}} refers to the predicted common concept and the \textbf{\textcolor{independent_concept_color}{blue box}} refers to the predicted independent concept. See section~\ref{sec:alignment} for details about each row.}
\label{fig:fig2}
\end{figure*}

\vspace{-0.5em}
In this section, we qualitatively cover four cases for summarizing the effects of the selected concept and the predicted concept and how these two relate to what the actual entity to be localized was. Note that our proxy loss is trained with the \emph{selected concept} as the ground truth and predicts the \emph{common concept} and \emph{independent concept}. In Figure~\ref{fig:fig2}, \emph{common concept}, \emph{independent concept} and \emph{selected concept} are denoted by red, blue and gray blocks respectively. For the remainder of this section we use the term \emph{true concept} to refer to the actual entity to be localized.
\newline
\textbf{First row: Correct grounding of phrase.} In cases where the \emph{selected concept} and all \emph{predicted concepts} coincide with the \emph{true concept} to be localized, the generated heatmap produces a good localization of the phrase. This is shown in the first row of Figure~\ref{fig:fig2} with the concept \emph{`headlight'} and \emph{`picture'}.
\newline
\textbf{Second row: Incorrect grounding due to wrong concept-selection.} In cases where the \emph{selected concept} is incorrect, \emph{ie.} it's not the same as the \emph{true concept}, even with the correct decoding, the localizations produced are wrong. For example in the second row of Figure~\ref{fig:fig2}, instead of selecting \emph{`building'} and \emph{`switch'}, the incorrect selection of \emph{`top'} and \emph{`wall'} leads to localization which is correct for the selected ground truth, but incorrect for the phrase.
\newline
\textbf{Third row: Incorrect grounding due to wrong concept-learning.} In these cases, the selected concept is correct but the decoder predicts incorrect common/independent concepts, due to which the final phrase grounding is affected. For example in the third row of Figure~\ref{fig:fig2}, even though \emph{`window'} and \emph{`tire'} are correct \emph{selected concepts}, the concept-learning inaccurately predicts \emph{`glass'} and \emph{`car/vehicle'} which in turn generates a localization corresponding to the \emph{predicted concept}.
\newline
\textbf{Fourth row: Incorrect grounding due to challenging phrase-image pairs.} Lastly, there are some cases where the entity to be localized is either ambiguous or simply too hard (due to a small size in the image or due to a complicated phrase structure). In these cases, the grounding is incorrect across the different possibilities of alignment of the aforementioned concept. For example in the fourth row of Figure~\ref{fig:fig2}, the concept \emph{`pole'} exists at multiple visual locations while in the other example, the concept \emph{`lighter'} occupies a very small space in the visual region.
\vspace{-0.5em}
\section{Additional outputs from our model}
\begin{figure*}[thpb]
\centering
\includegraphics[width=0.95\textwidth]{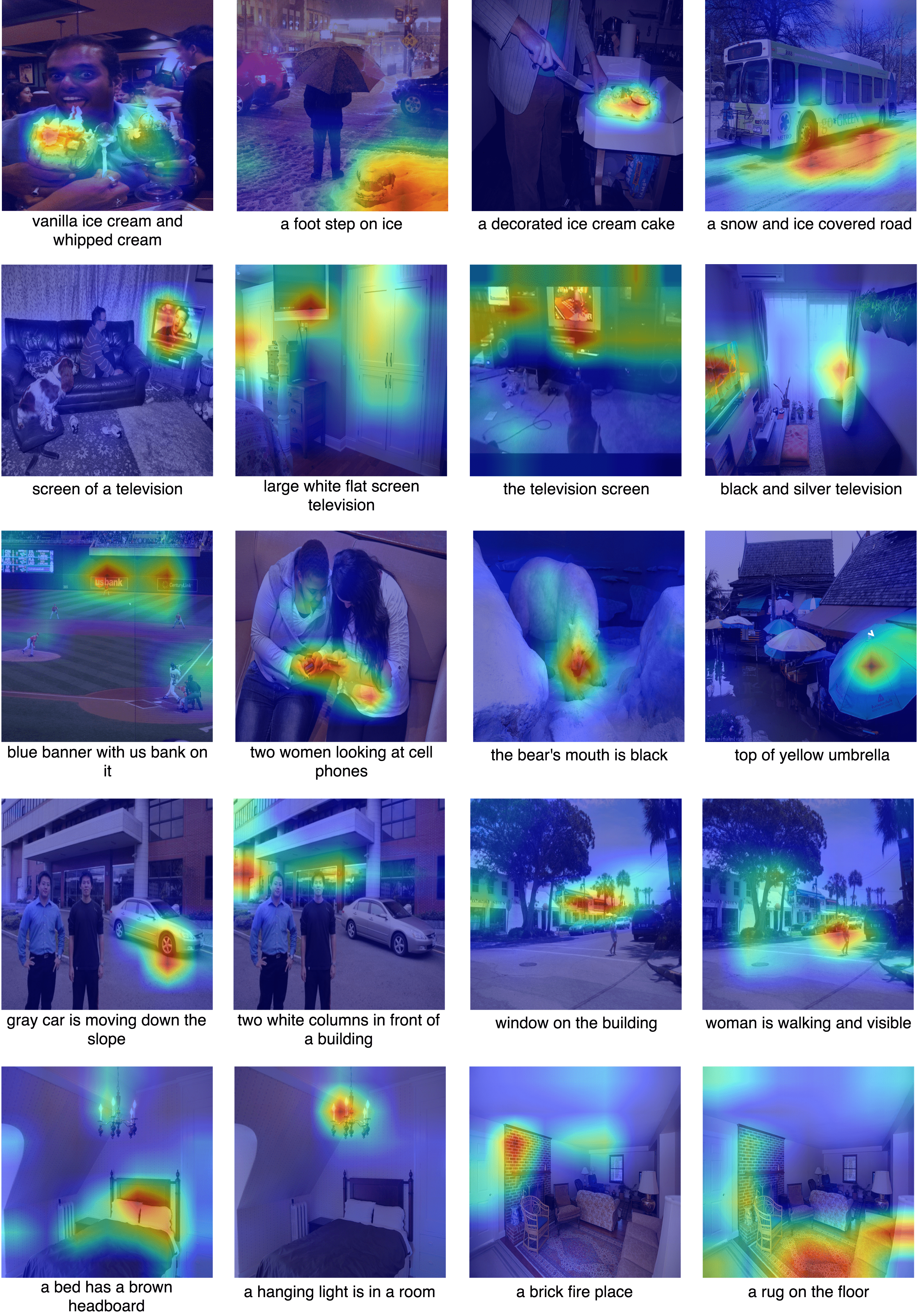}
\caption{Additional qualitative examples}
\label{fig:fig3}
\end{figure*}

\vspace{-0.5em}
In Figure~\ref{fig:fig3}, the first two rows show a typical concept batch with \emph{`ice'} and \emph{`television'} as the respective common concepts. The third row shows some small and challenging entities to be grounded. Finally, the fourth and fifth row highlight the ability of the model to output completely different heatmaps for the same image having differing phrases. The grounded heatmap appear to identify regularities like localizing \emph{`television'} towards the periphery near a wall or localizing \emph{`phone'} near the hands of a person. We also note that since the concept batch is trained with concepts from very diverse contexts, the model is forced to learn high-level semantics about the image (see first row).
\newpage

\newpage

\newpage

\end{document}